\documentclass[10pt,twocolumn,letterpaper]{article}

\usepackage{cvpr}              % To produce the CAMERA-READY version
\usepackage{graphicx}
\usepackage{amsmath}
\usepackage{amssymb}
\usepackage{booktabs}
\usepackage{bm}
\usepackage[noend]{algpseudocode}
\usepackage{algorithmicx,algorithm}
\usepackage{soul}
\usepackage{multicol}
\usepackage{multirow}
\usepackage{amsmath}
\usepackage{tabularx}

\usepackage{wrapfig}
\usepackage{pgfplots}
\usepgfplotslibrary{groupplots}

\usepackage{color}

\newlength\savewidth\newcommand\shline{\noalign{\global\savewidth\arrayrulewidth
  \global\arrayrulewidth 1pt}\hline\noalign{\global\arrayrulewidth\savewidth}}

\usepackage[pagebackref,breaklinks,colorlinks]{hyperref}

\usepackage[capitalize]{cleveref}
\crefname{section}{Sec.}{Secs.}
\Crefname{section}{Section}{Sections}
\Crefname{table}{Table}{Tables}
\crefname{table}{Tab.}{Tabs.}
\renewcommand\footnotemark{}

\def\cvprPaperID{*****} % *** Enter the CVPR Paper ID here
\def\confName{CVPR}
\def\confYear{2023}

\begin{document}

\title{CCD-3DR: Consistent Conditioning in Diffusion for Single-Image 3D Reconstruction}

\author{Yan Di$^{1}$, Chenyangguang Zhang$^{2}$, Pengyuan Wang$^{1}$, Guangyao Zhai$^{1}$, Ruida Zhang$^{2}$, \\ Fabian Manhardt$^{3}$, 
Benjamin Busam$^{1}$,
Xiangyang Ji$^{2}$, 
and Federico Tombari$^{1,3}$\\
\textsuperscript{1}Technical University of Munich, \textsuperscript{2}Tsinghua University,
\textsuperscript{3} Google,
\\
\tt\small{\{yan.di@, tombari@in.\}tum.de},
\tt\small{\{zcyg22, zhangrd21\}@mails.tsinghua.edu.cn}
\\
%\thanks{*Authors with equal contributions.}
%\thanks{Codes: \url{https://github.com/ZhangCYG/U-RED}}
}
\maketitle

%%%%%%%%% ABSTRACT
\begin{abstract}
   In this paper, we present a novel shape reconstruction method leveraging diffusion model to generate 3D sparse point cloud for the object captured in a single RGB image.
   Recent methods typically leverage global embedding or local projection-based features as the condition to guide the diffusion model.
   However, such strategies fail to consistently align the denoised point cloud with the given image, leading to unstable conditioning and inferior performance.
   In this paper, we present CCD-3DR, which exploits a novel centered diffusion probabilistic model for consistent local feature conditioning.
   We constrain the noise and sampled point cloud from the diffusion model into a subspace where the point cloud center remains unchanged during the forward diffusion process and reverse process.
   The stable point cloud center further serves as an anchor to align each point with its corresponding local projection-based features.
   Extensive experiments on synthetic benchmark ShapeNet-R2N2 demonstrate that CCD-3DR outperforms all competitors by a large margin, with over 40$\%$ improvement.
   We also provide results on real-world dataset Pix3D to thoroughly demonstrate the potential of CCD-3DR in real-world applications.
   Codes will be released soon.
   
\end{abstract}

%%%%%%%%% BODY TEXT
\begin{figure}[t]
    \centering
    \includegraphics[width=0.47\textwidth]{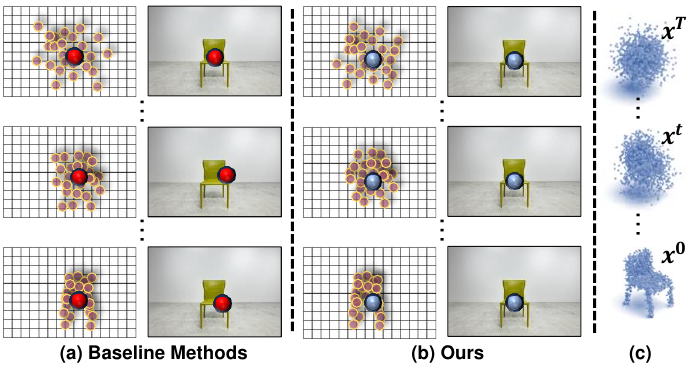}
    \caption{\textbf{Baseline Methods \textit{vs} Ours in Conditioning.}
    In reverse process (c), the generated point cloud is back-projected onto the feature map of the RGB image and local features are extracted around the projections.
    Fig~(a) and (b) compare the local feature extraction  methods of baseline methods~\cite{luo2021diffusion,melas2023pc2} and Ours. 
    In (a), during reverse process, the point cloud center gradually deviates (indicated by red points). 
    These deviations result in misaligned feature extraction, leading to a degradation in shape reconstruction quality.
    In contrast, our method, shown in (b), maintains the projection of the point cloud center unchanged throughout the reverse process (highlighted by blue points), serving as a stable anchor point that facilitates consistent extraction of local features.
    }
    \label{fig:teasor}
    \vspace{-0.2cm}
\end{figure}

\section{Introduction}
\label{sec:intro}
Single-image object reconstruction is a well-known ill-posed problem.
While deep learning methods have made remarkable strides in achieving high-quality reconstruction, further improvements are still necessary to meet the demands of real-world applications~\cite{zhai2023monograspnet,yang2019cubeslam}.
Recently, a new wave of methods leveraging \textbf{D}enoising \textbf{D}iffusion \textbf{P}robabilistic \textbf{M}odel (\textbf{DDPM})~\cite{ho2020denoising} has emerged~\cite{Cheng2022SDFusionM3,melas2023pc2,luo2021diffusion,melas2023realfusion,poole2022dreamfusion}, showcasing superior performance in various domains.
For single-image 3D reconstruction with diffusion models, DMPGen~\cite{luo2021diffusion} and PC$^2$~\cite{melas2023pc2} are two representative baselines.
In DMPGen, the condition is the global embedding of the target object, while in PC$^2$, in each step of the reverse process, the denoised point cloud is back-projected onto the feature map of the image to extract local feature for each point, which serves as the condition for the next reverse step.

However, directly applying diffusion models in single-image 3D reconstruction suffers from an inevitable challenge: uncontrollable center deviation of the point cloud.
Since each point inside the point cloud and predicted noise is independently modelled, under the single-image reconstruction setting, no geometric or contextual priors can be harnessed to control the point cloud center.
After each step of the reverse process in DDPM, the gravity center of the generated point cloud will be shifted slightly.
Therefore, from a random sampled Gaussian noise towards the target object, in reverse process, the center of the point cloud will continuously undergo disturbances until it reaches the center of the target object.
Based on our experimental findings, we have identified two problems caused by this center deviation.

\textbf{First}, the diffusion network needs to allocate capacity to handle the displacement of the point cloud center. 
It is crucial to ensure that the transition of the point cloud center from the initial Gaussian noise state to the final object reconstruction is appropriately managed. 
However, since the overall resource is limited, allocating network capacity to recover the center results in inferior performance in shape reconstruction.
\textbf{Second}, the center deviation causes misalignment and inconsistency in the local feature conditioning, as used in PC$^2$~\cite{melas2023pc2}. 
The misaligned feature adversely affects the subsequent denoising process in DDPM and degrades the overall quality of the final reconstruction.

To address the aforementioned problems, in this paper, we present CCD-3DR, which takes a single RGB image with corresponding camera pose as input and reconstructs the target object with sparse point cloud.
Instead of directly leveraging the off-the-shelf DDPM, we propose a novel \textbf{C}entered denoising \textbf{D}iffusion \textbf{P}robabilistic \textbf{M}odel (\textbf{CDPM}) that can enable consistent local feature conditioning in diffusion, which further significantly boosts the single-image reconstruction quality.
Our core idea is to constrain the added noise in diffusion process, the predicted noise and sampled point cloud in reverse process into a smaller subspace of the entire sampling space.
In this subspace, the center of corresponding noise or point cloud coincides with the origin throughout the diffusion and reverse processes.
Thereby, the point cloud center serves as an anchor in local feature extraction to align the point cloud with its corresponding projections consistently.

Based on CDPM, we design CCD-3D for single-image 3D object reconstruction.
In CCD-3D, to ensure that the noise and point cloud lie in the subspace defined in CDPM, a straightforward strategy is to iteratively generate samples in the entire space until one sample lie in the subspace.
However, this is time-consuming and infeasible in real implementations. 
We instead first sample in the entire space and then centralize the noise, predicted noise from the diffusion network, denoised point cloud in reverse process to send them to the subspace.
We follow PC$^2$~\cite{melas2023pc2} to back-project the point cloud onto the feature map of the image to extract local feature around each projection.

%-------------------------------------------------------------------------
In summary, our contributions are listed as follows,
\begin{itemize}
\setlength{\itemsep}{0pt}
\setlength{\parsep}{0pt}
\setlength{\parskip}{0pt}
\item 
We propose a novel centered denoising diffusion probabilistic model CDPM, which constrains the noise and point cloud in diffusion and reverse processes into a subspace where the point cloud center is forced to be coincide with the origin.
CDPM sacrifices some of DDPM's generation diversity in exchange for stability in the point cloud center.
\item 
We present a new single-image 3D object reconstruction pipeline CCD-3D, which leverages CDPM to consistently collect local features for the point cloud in diffusion, leading to superior performance in reconstruction quality. 
\item 
We evaluate CCD-3D on synthetic dataset ShapeNet-R2N2 to demonstrate its superiority over competitors. 
CCD-3D outperforms state-of-the-art methods by over $40\%$ under F-Score.
Additional experiments on real-world datasets demonstrate the potential of CCD-3D in real applications.
\end{itemize}

\section{Related Works}
 3D reconstruction of the object shape from a single image has been a research focus in the community\cite{kar2017learning,Wang2018Pixel2MeshG3, Wu2017MarrNet3S, Kar2014CategoryspecificOR, Li2018OptimizableOR, Li2018EfficientDP, Zhang2021ViewAwareGJ, Mao2021STDNetSA}. Although being an ill-posed problem, the shape priors in the large-scale training dataset can guide the reconstruction process with generalization ability.
 
\textbf{Non-Generative Reconstruction Models.}
 Early methods use 2D encoders~\cite{ronneberger2015u,he2016deep,simonyan2014very} to encode features and use 3D decoders~\cite{cciccek20163d,tran2015learning} to obtain shapes. The pioneering work such as 3D-R2N2  \cite{choy20163d} uses the occupancy grids as object shape representations and a following LSTM~\cite{hochreiter1997long} to fuse inputs from multiple views for prediction. The 2D features are extracted by a 2D CNN and projected to the 3D occupancy grids with a 3D deconvolutional neural network. LSM~\cite{kar2017learning} reprojects 2D features into voxel grids and decode shapes from these grids using a 3D convolutional GRU~\cite{cho2014learning}. Pix2Vox series \cite{xie2019pix2vox,xie2020pix2vox++} enjoy a serial architecture composing of a pretrained 2D CNN backbone and 3D transposed convolutional layers with multi-scale fusion to get better voxelized results. Since the voxel representations are limited by the resolution of voxel size, point cloud and mesh-based shape representations are favored to get rid of the limitation \cite{Hu2021SelfSupervised3M,Wang2020Pixel2Mesh3M,Zhang2018LearningTR,Henderson2019LearningS3,Erler2020Points2SurfLI,Mandikal2019Dense3P,Gkioxari2019MeshR,Wen2019Pixel2MeshM3,Pan2019DeepMR,Huang2023ShapeClipperS3}. More recent works utilizes implicit representations such as signed distance functions  \cite{Park2019DeepSDFLC,Xu2019DISNDI},  occupancy networks \cite{Mescheder2018OccupancyNL,Chen2018LearningIF} or neural radiance field for object shape generation \cite{Yu2020pixelNeRFNR, Wang2021NeuSLN,Jang2021CodeNeRFDN}.  Works such as AtlasNet~\cite{groueix2018papier} directly generates surface points and reconstructs object meshes, decomposing the problem into the assembly of multiple predicted patches. However, it proves that the auto-encoder architecture has limited generation ability, resulting in less diverse results. 

\textbf{Generative Reconstruction Models.}
Generative reconstruction models, in contrast to routines mentioned above, estimate the shape distribution in a more explicit way to generate plausible shapes.  For the first time to generate point clouds from single-view images, Fan et al. \cite{Fan_2017_CVPR} builds a point cloud generation network upon variational autoencoders (VAEs)~\cite{kingma2013auto} to generate multiple plausible shapes. By incorporating both VAEs and generative adversarial networks (GANs)~\cite{goodfellow2014generative}, 3D-VAE-GAN \cite{Wu2016LearningAP} samples latent codes from a single-view image as the condition and outputs 3D shapes through 3D GAN generators. However, It heavily relies on class labels for reconstruction. 3D-aware GANs such as StyleSDF \cite{OrEl2021StyleSDFH3} and Get3D \cite{Gao2022GET3DAG} can simultaneously synthesize 2D images and 3D detailed meshes. However, these methods suffer from instabilities and mode collapse of GAN training. 
%As another kind of generative models, Normalzing Flow [] inspires PointFlow \cite{Yang2019PointFlow3P} to propose a normalized point flow to guide the point cloud generalization process from shape priors. 

Recently, diffusion models \cite{song2019generative,song2020improved,ho2020denoising} exhibit advanced generation ability in such as text-to-image \cite{Rombach2021HighResolutionIS}, text-to-shape \cite{Nichol2022PointEAS} areas, enjoying more stable training phase and elegant mathematical explainability. Thereby, various point cloud related tasks take advantage of diffusion models to get results in higher quality. Luo et al. \cite{luo2021diffusion} firstly applies the diffusion process in the point cloud generation task. LION \cite{Zeng2022LIONLP} further generalizes the point cloud in the hierarchical latent space with diffusion. Similarly, Lyu et al. \cite{Lyu2021ACP} utilizes the point diffusion for shape completion. Point-Voxel Diffusion \cite{Zhou_2021_ICCV} combines multiple representations in the diffusion process to generate stabel results. To get the texture information for the point cloud, \cite{Nichol2022PointEAS} generates colored point clouds as the diffusion output for better visualization. Theoretically, such methodology can be readily leveraged into the single-view reconstruction task by regarding the RGB information as the condition \cite{poole2022dreamfusion, melas2023pc2}. Most recent method PC$^2$ \cite{melas2023pc2} projects point clouds in the reverse diffusion process to image plane to query 2D features as shape and color conditions.

\begin{figure*}[t]
    \centering
    \includegraphics[width=0.99\textwidth]{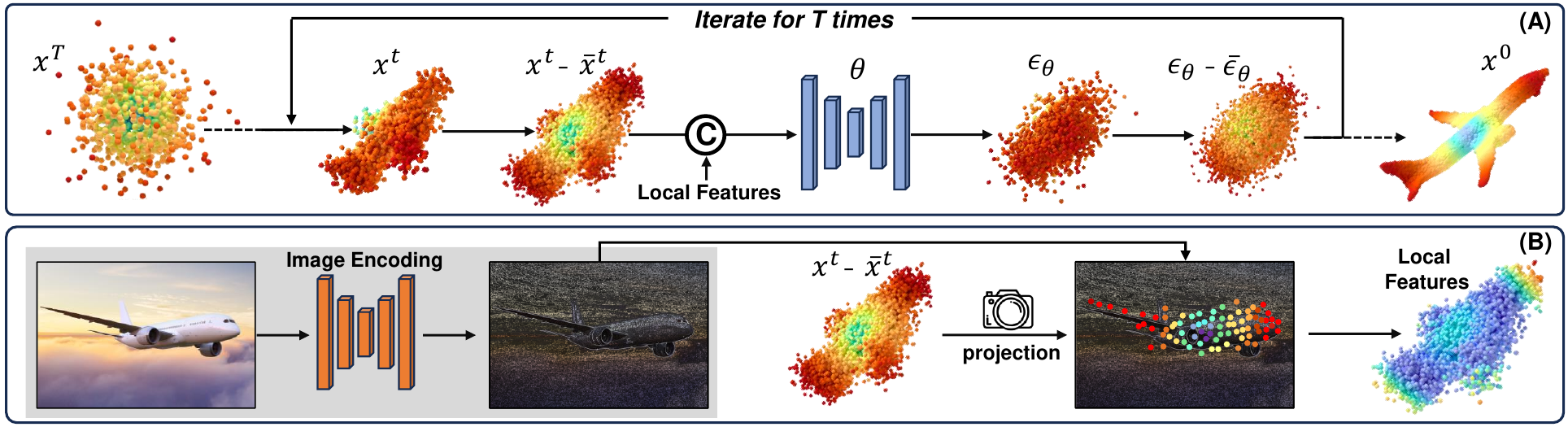}
    \caption{\textbf{Architecture of CCD-3D.} 
    Given a single RGB image (capturing the airplane) as the input, CCD-3D aims to reconstruct the object with CDPM.
    We first leverage a pre-trained MAE~\cite{he2022masked} model to extract feature maps from the image and interpolate them to the same size as the image (shown in the grey block).
    The feature maps are utilized to provide local conditions for each point in the denoised point cloud.
    Block (A) demonstrates the reverse process of CDPM. 
    At step $t$, point cloud $x^t$ is first centralized to $x^t-\bar{x}^t$ and then concatenated with the local features.
    The U-Net denoiser $\theta$ predicts noise $\epsilon_\theta$ and centralize it with $\epsilon_\theta - \bar{\epsilon}_\theta$.
    The point cloud $x^{t-1}$ can finally be recovered using Eq.~\ref{qxtxt_1}.
    Block (B) shows the local feature extraction process.
    We back-project the centered point cloud $x^t-\bar{x}^t$ onto the image using the given camera pose and collect features around the projections to serve as the local features. 
    }
    \label{fig:pipeline}
\end{figure*}

\section{Method}
In the following sections, we outline our methodology. 
We commence by providing a brief overview of point diffusion models, laying the groundwork for our approach. 
Subsequently, we elucidate the enhancements we have made to the traditional DDPM, with the intention of augmenting its effectiveness in the realm of single-image reconstruction. 
These adaptations result in our innovative centered denoising diffusion probabilistic model (CDPM). 
Lastly, we provide a comprehensive explanation of our single-image reconstruction pipeline CCD-3DR, which is constructed based on CDPM.

\subsection{Preliminaries: Diffusion Models}
Diffusion denoising probabilistic models are a class of generative models inspired by non-equilibrium thermodynamics.
It can iteratively move a set of Gaussian noise towards a uniform and clean point cloud capturing the target object.
DDPM contains two Markov chains called the diffusion process and the reverse process.
The two processes share a length of $T=1K$ steps.

\textbf{Diffusion Process.}
Let $p_{0}$ be the potential distribution of the complete object point cloud $x$ in the dataset and $p_{T}$ be the standard Gaussian distribution $p_{T} \sim \mathcal{N}(0_{3N}, I_{3N \times 3N})$,  
The diffusion process iteratively adds Gaussian noise $\epsilon$ into the clean data distribution $p_{0}$ according to the Markov Chain Rule until $p_{0}$ reaches $p_{T}$.
Formally, let $x^{0} \sim p_{0}$, then
\begin{equation}
    \label{diff_process}
    \centering
    \begin{split}
     &q(x^{1:T}|x^{0})=\prod_{t=1}^{T}q(x^{t}|x^{t-1}),\\
     where \quad &q(x^t|x^{t-1})=\mathcal{N}(x^t;\sqrt{1-\beta_t}x^{t-1}, \beta_t\bm{I}).   
    \end{split}
\end{equation}
\noindent The hyperparameter $\beta_t$ is pre-defined small constants.
We use the subscript to denote the diffusion step $t$.
Each $q(x^{t}|x^{t-1})$ is a Gaussian distribution and $q(x^t|x^0)$ can be reparameterized as,
\begin{equation}
    \label{qxtx0}
    q(x^t|x^0) = \sqrt{\bar{\alpha}_t}x^0+\epsilon \sqrt{1-\bar{\alpha}_t},
\end{equation}
where $\alpha_t=1-\beta_t$, $\bar{\alpha}_t=\prod_{s=0}^{t}\alpha_s$, and $\epsilon \sim \mathcal{N}(0, \bm{I})$.

From Eq.~\ref{qxtx0}, for point diffusion, we can infer that if $x^0$ is sampled from a zero-mean distribution $p_{0}$, considering $\epsilon$ is also zero-mean, $q(x^t|x^0)$ can be modelled as a zero-mean distribution, which mean for any $t \in [0, T]$, the diffusion process will generate a zero-mean distribution at this step.
In this paper, we utilize this derivation to boost single-image 3D reconstruction.

\textbf{Reverse Chain.}
The reverse process is also a Markov process that removes the noise added in the diffusion process.
In this paper, the reverse process is conditioned on the conditioner, an RGB image $I$ capturing the object.
We start with a sample $x^{T} \sim p_T$, and then iteratively sample from $q(x^{t-1}|x^{t}, f(I))$, where $f(I)$ denotes features extracted from $I$ to incorporate local or global supervision into the reverse process.
When the sampling step is sufficiently large ($T$ is large), $q(x^{t-1}|x^{t}, f(I))$ can be well approximated with an isotropic Gaussian distribution with a fixed small covariance,
\begin{equation}
    \label{qxtxt_1}
    \begin{split}
      &q(x^{t-1}|x^t, f(I))=\mathcal{N}(x^{t-1};\mu_\theta(x^t, f(I)),\sigma_t^2I), \\
      &\mu_\theta(x^t, f(I))=\frac{1}{\sqrt{\alpha_t}}(x^t-\frac{\beta_t}{\sqrt{1-\bar{\alpha}_t}}\epsilon _\theta(x^t, f(I))), 
    \end{split}
\end{equation}
where $\mu_\theta$ is the estimated mean.
Thus we can use the network parameterized by $\theta$ to directly learn $\epsilon_\theta$ under the condition $f(I)$.

\textbf{DDPM-Based Reconstruction}
Consider a 3D point cloud with $N$ points, DDPM-based reconstruction methods~\cite{luo2021diffusion,melas2023pc2} learn a diffusion model $S_\theta: \mathbb{R}^{3N} \rightarrow \mathbb{R}^{3N}$ to denoise the randomly sampled point cloud from $p_{T}$ into a recognizable object from target distribution $p_{0}$.
Specifically, at each step $t$, the noise is predicted as the offset of each point from the current coordinate in $x^{t}$ to $x^{t-1}\sim q(x^{t-1}|x^{t}, f(I))$.
Then we sample from $q(x^{t-1}|x^{t}, f(I))$ to obtain $x^{t-1}$.
As for conditioning, DPMGen~\cite{luo2021diffusion} encodes the given RGB image into a single global latent vector $z$ and concatenate $z$ with obtained point cloud at each step during the reverse process.
PC$^2$~\cite{melas2023pc2} goes one step further by introducing local point-wise features for fine-grained geometry cues.
It updates the local feature of each point at each step $t$ by backprojecting the point cloud $x^t$ onto the feature map using the given camera extrinsic $[R_c|T_c]$ and perspective projection matrix $\pi_c$,
\begin{equation}
    \label{projection}
    Proj(x^t) = \pi_c(R_cx^t+T_c).
\end{equation}
Then local features $f(I)$ around the projections $Proj(x^t)$ are aggregated with rasterization.
These two methods~\cite{luo2021diffusion,melas2023pc2} are selected as our baselines in this paper.

\subsection{Bottlenecks in DDPM Reconstruction}
We now analyze the limitations of directly applying DDPM in 3D reconstruction like in DPMGen and PC$^2$~\cite{luo2021diffusion,melas2023pc2}.
There are two bottlenecks that deteriorate the performance of these methods.

\textbf{First}, predicting the center bias is challenging for the network in the reverse process.
Since we assume the variances are constant in all Gaussian distributions, we only need to analyze the center of each denoised point cloud.
From $x^t$ to $x^{t-1}$, in Eq.~\ref{diff_process} and~\ref{qxtxt_1}, we have,
\begin{equation}
    E(\bar{x}^{t-1}) = \frac{1}{\sqrt{\alpha_t}}E(\bar{x}^t), \quad E(\bar{\epsilon} _\theta(x^t, f(I)))=0.
\end{equation}
Thus after sampling for $x^{t-1}$, we can obtain,
\begin{equation}
    \label{bar}
    \bar{x}^{t-1} = \frac{1}{\sqrt{\alpha_t}}(\bar{x}^t-\frac{\beta_t}{\sqrt{1-\bar{\alpha}_t}}\bar{\epsilon} _\theta(x^t, f(I))) + \Delta_t,
\end{equation}
where $\Delta_t$ is center bias generated by random sampling from Gaussian distribution for $x^{t-1}$.
When $\bar{x}^{T} \neq \bar{x}^{0}$, the network $\theta$ needs to move the center of the denoised point cloud from $\bar{x}^{T}$ towards $\bar{x}^{0}$ under the following handicaps.
First, $E(\bar{\epsilon} _\theta(x^t, f(I)))=0$, while the network needs to predict non-zero-mean noise $\epsilon$ in several steps to move $\bar{x}^{T} \rightarrow \bar{x}^{0}$.
Second, the network needs to overcome $\Delta_t$.
Last, each point in $x^{T:0}$ is independently modelled in diffusion, and no constrains are incorporated to control the development of the point cloud center.
Experiments demonstrate that accurately recovering $\bar{x}^{0}$ is a very hard job for the network.
And wasting network capacity in recovering center also results in poor performance in shape reconstruction.

\iffalse
As illustrated in Eq.~\ref{qxtx0}, we sample the point cloud $x^t$ with $N$ points from $q(x^t|x^0)$, which is a combination of $x^0$ and a Gaussian noise $\epsilon$.
If $N \rightarrow \infty$, then the sampled noise $\epsilon^t$ from $\epsilon$ will be strictly zero-mean.
Therefore, the center of $x^0$ will gradually approach $\bm{0}$ in the diffusion process, under the balance of $\sqrt{\bar{\alpha}_t}$ and $\sqrt{1-\bar{\alpha}_t}$，
\begin{equation}
    
\end{equation}

However, in real implementation, due to limited network processing capability, we only sample a sparse set of points ($N=8192$), which makes the added noise in diffusion process and the predicted noise in reverse process contain undesired small center deviations.
For diffusion process, as derived from Eq.~\ref{qxtx0} from step $0$-$t$, the accumulated center bias can be represented as,
\begin{equation}
    \label{differror}
    \Delta \bar{x}^t = \bar{x}_t - s(x^t)= \Delta s(x^t) + \sqrt{1-\bar{\alpha}_t}\Delta\epsilon 
\end{equation}
where $s(x^t)$ denotes sampled $x^t$ from the distribution $q(x^t|x^0)$.
$\Delta s(x^t)$ and $\Delta\epsilon$ denote errors from sampling of point cloud and noise respectively.
In reverse process, derived from Eq.~\ref{qxtxt_1}, the center bias from 

This further mislead the reconstruction of $x^{0}$ as the deviations accumulate from $T \rightarrow 0$.
\fi

\textbf{Second}, the change of the point cloud center makes the local feature conditioning inconsistent.
As in PC$^2$, the difference $\Delta_{Proj}$ in projections of $Proj(\bar{x}^{t-1})$ and $Proj(\bar{x}^{t})$ can be derived as
\begin{equation}
    \Delta_{Proj} = \pi_c(R_c(\bar{x}^{t-1} - \bar{x}^t)+T_c).
\end{equation}
If $\Delta_{Proj}$ is sufficiently large, the features collected for the point center can be totally different from $x^{t}$ to $x^{t-1}$, which will mislead the following denoising steps.
Moreover, since we only use single RGB image as conditioner, we have no contextual or geometric constraints to rectify this misalignment.

\begin{algorithm}[t]
\caption{CDPM: Training} 
\label{algorithm1}
{\bf Repeat:} 
\begin{algorithmic}[1]
\State $x^0 \sim q(x^0)$, \quad $x^0=x^0-\bar{x}^0$
\State $t \sim$ Uniform($\{1, 2,..., T\}$)
\State $\epsilon \sim \mathcal{N}(\bm{0}, \bm{I})$, $\epsilon=\epsilon-\bar{\epsilon}$
\State Take gradient descent step on: 
$\bigtriangledown_\theta\left\|\epsilon - \epsilon_\theta(x^t, f(I)) \right\|^2$
\end{algorithmic}
{\bf Until} converged
\end{algorithm}
\vspace{-0.3cm}

\begin{algorithm}[t]
\caption{CDPM: Sampling}
\label{algorithm2}
\begin{algorithmic}[1]
\State $x^T \sim \mathcal{N}(\bm{0}, \bm{I})$, \quad $x^T=x^T-\bar{x}^T$
\For{t=T,...,1}
    \State $\epsilon_\theta=\epsilon_\theta-\bar{\epsilon}_\theta$
    \State $x^{t-1} \sim q(x^{t-1}|x^{t})$, $x^{t-1}=x^{t-1}-\bar{x}^{t-1}$
\EndFor
\Return $x^0$
\end{algorithmic}
\end{algorithm}

\subsection{From DDPM to CDPM}
To address the aforementioned bottlenecks, we propose a novel CDPM model that is designed for single-view 3D reconstruction.
The core idea of CDPM is straightforward.
To eliminate the influence of center bias in reverse process, we add the following constraint,
\begin{equation}
    \label{constraints}
    \bar{x}^t=\bm{0}, \quad t = 0,1,2...,T.
\end{equation}
This constraint enforces that the denoised point cloud in each step to be zero-mean, so that the center remains unchanged during the reverse process.
As shown in Eq.~\ref{qxtx0} and Eq.~\ref{qxtxt_1}, if Eq.~\ref{constraints} holds, we have $\bar{\epsilon} = \bm{0}$, $\bar{\epsilon} _\theta(x^t, f(I))=\bm{0}$.
Let $\mathbb{S}_{x^t}$ denote the space of all possible samplings from the distribution $q(x^t|x^{t+1})$, then the space under the constraint Eq.~\ref{constraints} $\mathbb{S}_{x^t, \bar{x}^t=\bm{0}}$ is a subspace of $\mathbb{S}_{x^t}$, \textit{i.e.} $\mathbb{S}_{x^t, \bar{x}^t=\bm{0}} \subset \mathbb{S}_{x^t}$.
Similarly, we define $\mathbb{S}_{\epsilon}$, $\mathbb{S}_{\epsilon, \bar{\epsilon}=0}$, $\mathbb{S}_{\epsilon_\theta}$, $\mathbb{S}_{\epsilon_\theta, \bar{\epsilon_\theta}=0}$.
In summary, from DDPM to CDPM, we constrain $x^t, \epsilon, \epsilon_\theta$ in a smaller subspace,
\begin{equation}
    \label{changes}
    \begin{split}
     &\textit{DDPM}: x^t \in \mathbb{S}_{x^t}, \epsilon \in \mathbb{S}_{\epsilon}, \epsilon_\theta \in \mathbb{S}_{\epsilon_\theta} \Longrightarrow \\
     &\textit{CDPM}: x^t \in \mathbb{S}_{x^t, \bar{x}^t=\bm{0}}, \epsilon \in \mathbb{S}_{\epsilon, \bar{\epsilon}=0}, \epsilon_\theta \in \mathbb{S}_{\epsilon_\theta, \bar{\epsilon_\theta}=0}.
    \end{split}
\end{equation}

Therefore, we prioritize the stability of the point cloud center to a certain extent, sacrificing a portion of the diversity in diffusion models. 
For point cloud $x^t$ in reverse process, after obtaining $q(x^{t}|x^{t+1})$, we can sample multiple times until the sampled point cloud lie in $\mathbb{S}_{x^t, \bar{x}^t=\bm{0}}$.
However, such strategy is infeasible in real implementation.
Thereby we simply first sample in $\mathbb{S}_{x^t}$ and then centralize the point cloud to project it into $\mathbb{S}_{x^t, \bar{x}^t=\bm{0}}$, so as to $\epsilon$ and $\epsilon_\theta$.

Specifically, as explained in Alg.~\ref{algorithm1} and Alg.~\ref{algorithm2}, we first build a dataset composed of $M$ data pairs $\mathcal{D}=\{(x_i, I_i)| 1\leq i\leq M\}$, where $x_i$ denotes the $i$th ground truth point cloud sampled from the object mesh, and $I_i$ is the corresponding RGB image capturing the object.
Compared to DDPM, CDPM mainly made improvements in 3 points.

\textbf{First}, the point clouds in $\mathcal{D}$ are centralized as $x_i = x_i - \bar{x}_i$, where $\bar{x}_i$ denotes the gravity center of $x_i$, establishing a new zero-mean dataset $\bar{\mathcal{D}}={(\bar{x}_i, I_i)}$.

\textbf{Second}, for noise $\epsilon$ added in diffusion process for training and the noise $\hat{\epsilon}$ predicted in reverse process, we also centralize them as  $\epsilon = \epsilon - \bar{\epsilon}$ and $\hat{\epsilon} = \hat{\epsilon} - \bar{\hat{\epsilon}}$, where $\bar{\epsilon}$ and $\bar{\hat{\epsilon}}$ denote the corresponding gravity centers.

\textbf{Third}, during inference, for $x^{t-1}$ sampled from $q(x^{t-1}|x^{t}, f(I))$, we also centralize $x^{t-1}$ with $x^{t-1} = x^{t-1} - \bar{x}^{t-1}$.
From Eq.~\ref{qxtx0}, since we keep $x^{0}$ and $\epsilon$ to be zero-mean, the diffused point cloud in each step $t$ should be zero-mean.

The advantages of CDPM over DDPM in single-image reconstruction can be summarized as follows.

\textbf{First}, our zero-mean reverse process starts with a zero-mean Gaussian noise and arrives at the zero-mean reconstruction $x^{0}$ after $T$-step zero-mean denoising.
This zero-mean nature of the reverse process provides a useful regularization for the network to focus more on the object shape rather than tracking the center of the point cloud.
Therefore, our CDPM outperforms the previous DDPM-reconstruction methods even with only global embedding of the object like in~\cite{luo2021diffusion}.

\textbf{Second}, CDPM enables consistent local feature conditioning in the reverse diffusion process.
As in PC$^2$~\cite{melas2023pc2}, the point cloud is back-projected onto the image feature map to extract local point-wise feature as conditioning.
However, due to the uncontrollable center bias in reverse process, the projection of each point may gradually deviate, making the local feature aggregation process fail and further deteriorating the final reconstruction quality.
In contrast to DDPM-based PC$^2$, our method CDPM keeps the gravity center of the denoised point cloud in each step to coincide with the origin, which further serves as an anchor point in local feature collection.
The projection of this anchor point remains the same in the reverse process, and thus align the point cloud with the feature map to obtain consistent features.

\subsection{CCD-3DR}
For fair comparison with baseline methods, we follow PC$^2$~\cite{melas2023pc2} to use MAE~\cite{he2022masked} to extract 2D feature maps from the given RGB image.
The feature maps are of equal length and width of the input image to facilitate point cloud projection.
For the diffusion network $\theta$ used to predict the noise $\epsilon_\theta$, we adopt the Point-Voxel CNN (PVCNN) ~\cite{liu2019pvcnn}.
We use the classic $\mathcal{L}_2$ loss to supervise the training of $\theta$, as specified in Alg.~\ref{algorithm1}.

\section{Experiments}
\noindent\textbf{Datasets}.
We evaluate CCD-3DR on the synthetic dataset ShapeNet-R2N2~\cite{choy20163d,chang2015shapenet} and real-world dataset Pix3D~\cite{sun2018pix3d}.
ShapeNet contains a diverse collection of 3D models spanning various object categories, such as furniture, vehicles, and more. 
The dataset is meticulously annotated, providing not only the 3D geometry of the objects but also rich semantic information, making it an essential tool for quantitative evaluation of single-view reconstruction methods.
We follow baseline methods~\cite{melas2023pc2, yagubbayli2021legoformer, xie2020pix2vox++} to use the R2N2~\cite{choy20163d} subset along with the official image renderings, train-test splits, camera intrinsic and extrinsic matrices.
The R2N2 subset covers 13 categories in total. 
Pix3D~\cite{sun2018pix3d} is a large-scale benchmark of diverse image-shape pairs with pixel-level 2D-3D alignment.
Previous methods~\cite{Cheng2022SDFusionM3, xie2019pix2vox,xie2020pix2vox++, sun2018pix3d} only harness the \textit{chair} category and exclude the occluded samples.
Since our method needs to use all data to demonstrate robustness towards occlusion, we leverage 3 categories: $\{$\textit{chair}, \textit{table}, \textit{sofa}$\}$ and randomly generate train-test split with about 90$\%$ samples as the training set and the remaining as the testing set.
Details are provided in the Supplemetary Material.

\noindent\textbf{Implementation Details.}
We implement CCD-3DR in PyTorch and evaluate the method on a single GeForce RTX 3090Ti GPU with 24GB memory.
For ShapeNet-R2N2~\cite{choy20163d,chang2015shapenet}, we first resize the provided images of size $137\times137$ to $224\times224$ and adjust the focal length accordingly.
We follow prior work to use 8192 points in training and inference for fairness in computing the F-Score.
On Pix3D~\cite{sun2018pix3d}, since the images are of different sizes, we first crop the image with the given bounding box to obtain an object-centric image and then resize it to $224\times224$. 
The camera intrinsic matrix is also adjusted correspondingly.
During training, we train CCD-3DR with batch size 16 for 100K steps in total, following PC$^2$~\cite{melas2023pc2}.
We use the AdamW optimizer with a dynamic learning rate with warmup which increases from $1\times10^{-5}$ to $1\times10^{-3}$ in the first 2K steps, and then decays exponentially until $0$ in the following 98K steps.

\noindent\textbf{Baselines.}
We select DDPM-based DMPGen~\cite{luo2021diffusion} and PC$^2$~\cite{melas2023pc2} as our baseline methods.
On ShapeNet-R2N2, we compare with the official results of PC$^2$.
Since DMPGen doesn't provide results of single-view reconstruction on ShapeNet-R2N2, we reimplement it by using pre-trained MAE~\cite{he2022masked} to extract global shape code, and then follow the diffusion process in the original paper to reconstruct the object, denoted as DMPGen$^{*}$.
We provide three variants of CCD-3DR on ShapeNet-R2N2, in which \textit{Ours} uses only local features like in PC$^2$, \textit{Ours-G} leverages only global features as DMPGen$^{*}$ and \textit{Ours-(G+L)} incorporates both local and global features for reconstruction, as shown in Tab.~\ref{tab-gl_ab}.
On Pix3D, we retrain PC$^2$ and DMPGen$^{*}$ under the same settings of CCD-3DR.

\noindent\textbf{Evaluation Metrics}.
We use Chamfer Distance (CD) and F-Score@$0.01$~\cite{melas2023pc2, Cheng2022SDFusionM3} as the evaluation metrics.
CD quantifies the dissimilarity between two sets of points by measuring the minimum distance from each point in one set to its nearest point in the other set.
To compensate the problem that CD can be sensitive to outliers, we also report F-Score with the threshold 0.01, \textit{i.e.}, for each reconstructed point, if its nearest distance to the ground truth point cloud lies below the threshold, it is considered as correctly predicted.
Note that previous methods~\cite{choy20163d, yagubbayli2021legoformer, xie2020pix2vox++} typically report the results using the voxelized $32^{3}$ volume as the shape representation, which quantizes the sampled points and fails to reflect the reconstruction quality of fine-grained structures.
Therefore, we follow PC$^{2}$~\cite{melas2023pc2} to use sampled points from the object mesh as the ground truth.
Results of other methods~\cite{choy20163d, yagubbayli2021legoformer, xie2020pix2vox++} are re-evaluated using the same setting for fair comparisons.

\begin{table*}[t!]
\centering
\scalebox{0.99}{
\begin{tabular}{cccc|ccc||ccc}
\shline
\multirow{2}{*}{Category} & 3D-R2N2 &
LegoFormer & Pix2vox++ & DMPGen$^{*}$ & PC$^2$ & \multirow{2}{*}{Ours} &
DMPGen$^{*}$ &PC$^{2}$ &Ours \\
 & \cite{choy20163d} & \cite{yagubbayli2021legoformer} & \cite{xie2020pix2vox++} & \cite{luo2021diffusion} & \cite{melas2023pc2} &  & Oracle & Oracle & Oracle \\
\hline
airplane & 0.225 & 0.215 & 0.266 &0.454& 0.473 & \textbf{0.725}& 0.565 & 0.681& \textbf{0.785}\\
bench & 0.198 & 0.241 & 0.266 & 0.175& 0.305 & \textbf{0.480}&0.289 & 0.444 & \textbf{0.573}\\
cabinet & 0.256 & 0.308 & \textbf{0.317} & 0.087& 0.203 & 0.282& 0.111& 0.303&  \textbf{0.371}\\
car & 0.211 & 0.220 & 0.268 &0.310& 0.359 & \textbf{0.395}& 0.402& 0.420& \textbf{0.466}\\
chair & 0.194 & 0.217 & 0.246 &0.171& 0.290 & \textbf{0.335}& 0.312& 0.377& \textbf{0.406}\\
display & 0.196 & 0.261 & 0.279 &0.211& 0.232 & \textbf{0.381}& 0.236 & 0.357& \textbf{0.487}\\
lamp & 0.186 & 0.220 & 0.242 &0.207& 0.300 & \textbf{0.438}& 0.347 & 0.399& \textbf{0.490}\\
loudspeaker & 0.229 & 0.286 & \textbf{0.297} &0.113& 0.204 & 0.219&0.126 & 0.288&  \textbf{0.291}\\
rifle & 0.356 & 0.364 & 0.410 &0.474& 0.522 & \textbf{0.762}& 0.663& 0.686& \textbf{0.828}\\
sofa & 0.208 & 0.260 & 0.277 &0.078& 0.205 & \textbf{0.293}& 0.106& 0.298& \textbf{0.349} \\
table & 0.263 & 0.305 & 0.327 &0.155& 0.270 & \textbf{0.427}& 0.310& 0.420& \textbf{0.488}\\
telephone & 0.407 & 0.575 & \textbf{0.582} &0.333& 0.331 & 0.423& 0.464& 0.523& \textbf{0.598}\\
watercraft & 0.240 & 0.283 & 0.316 &0.201& 0.324 & \textbf{0.475}& 0.399 & 0.424& \textbf{0.610}\\
\hline
Average & 0.244 & 0.289 & 0.315 &0.228& 0.309 & \textbf{0.433}& 0.333& 0.432& \textbf{0.519}\\
\shline 
\end{tabular} }
\vspace{-0.3cm}
\caption{\textbf{Performance on ShapeNet-R2N2.} 
We compare our method with competitors under F-Score@0.01. 
The Oracle setting refers to predicting 5 samples of each image and selecting the best prediction as the final result.
}
\label{tab-shapenet}
\end{table*}

\subsection{Comparisons with State-of-the-Art Methods.}

\noindent\textbf{Performance on Synthetic Dataset ShapeNet-R2N2.}
In Tab.~\ref{tab-shapenet}, we compare CCD-3DR with state-of-the-art competitors on ShapeNet-R2N2 under the F-Score@0.01 metric.
3D-R2N2~\cite{choy20163d}, Legoformer~\cite{yagubbayli2021legoformer}, Pix2vox++~\cite{xie2020pix2vox++} are voxel-based methods, while DMPGen~\cite{luo2021diffusion}, PC$^2$~\cite{melas2023pc2} are diffusion-based methods, serving as baselines of CCD-3DR.
From Tab.~\ref{tab-shapenet}, it can be clearly deduced that our method CCD-3DR achieves state-of-the-art performance in 10 out of 13 categories.
Considering the \textit{Average} performance, CCD-3DR outperforms previous best method Pix2vox++ with $0.433$ \textit{vs.} $0.315$, about a $37.5\%$ leap forward.
Furthermore, comparing with diffusion-based baseline method PC$^2$, CCD-3DR demonstrates superior performance under all the categories, and improves PC$^2$ by $40.1\%$, with $0.433$ \textit{vs.} $0.309$.
We also report the \textbf{Oracle} results, following the setting in PC$^2$, where for each test image, we predict 5 possible reconstruction results and select the one with highest F-Score@0.01 as the final result.
Under the \textbf{Oracle} setting, our method surpasses all competitors by a large margin, with about a $20.1\%$ improvement over PC$^2$ Oracle.

\begin{figure}[t]
    \centering
    \includegraphics[width=0.48\textwidth]{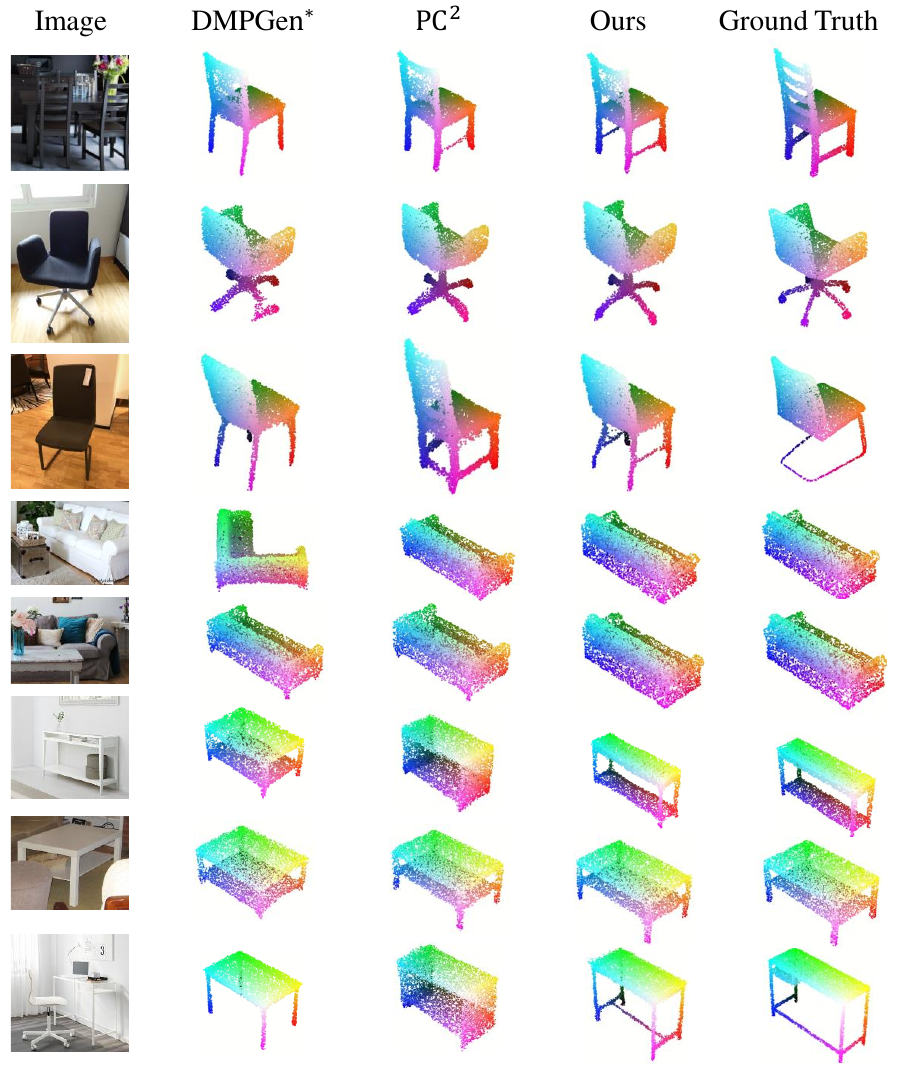}
    \caption{\textbf{Qualitative comparisons on real-world dataset Pix3D}~\cite{sun2018pix3d}. We compare with diffusion-based baseline methods DMPGen$^{*}$~\cite{luo2021diffusion} and PC$^{2}$~\cite{melas2023pc2}.}
    \label{fig:exp_pix3d}
\end{figure}

\begin{figure}[t]
    \centering
    \includegraphics[width=0.48\textwidth]{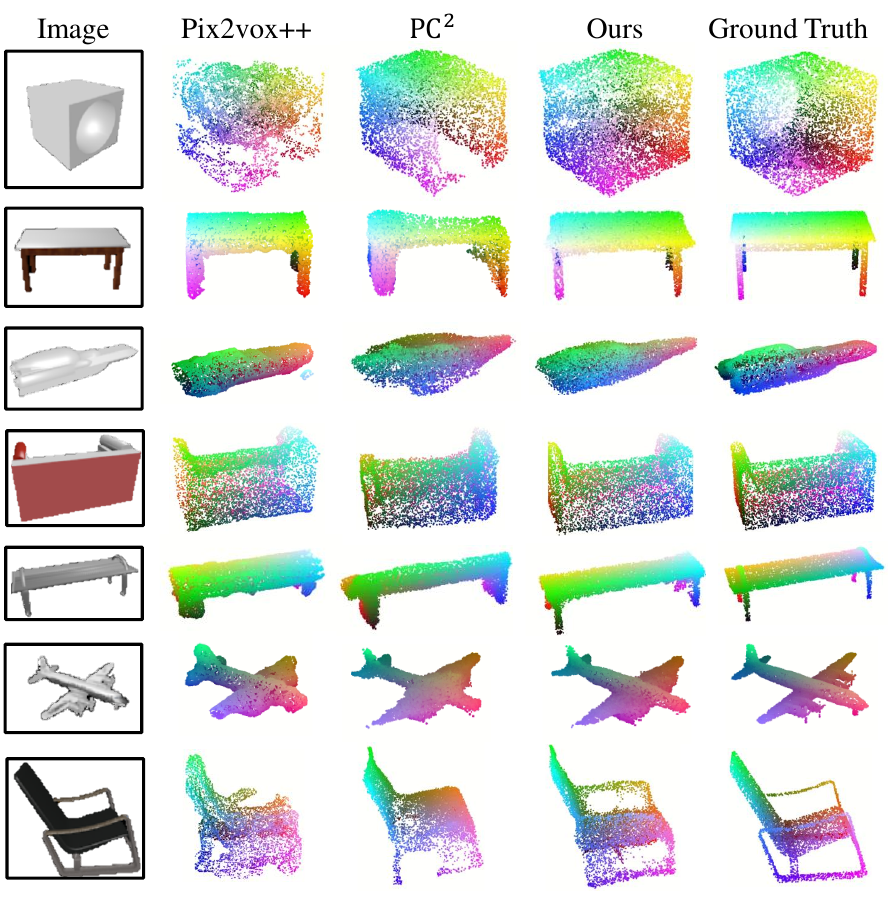}
    \caption{\textbf{Qualitative comparisons on synthetic dataset ShapeNet-R2N2}~\cite{choy20163d,chang2015shapenet}.
    Our method can recover fine-grained structures accurately, like the handle of the chair.}
    \label{fig:exp_shapenet}
\end{figure}

\begin{figure}[t]
    \centering
    \includegraphics[width=0.48\textwidth]{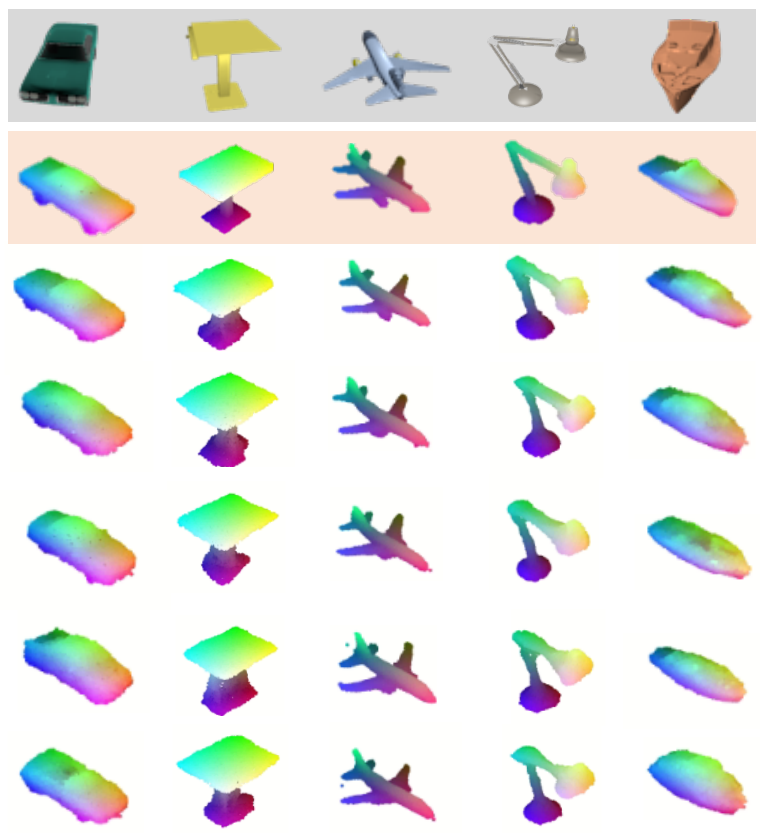}
    \caption{\textbf{Visualization of our Oracle results.}
    The input images and corresponding ground truth shapes are are highlighted with grey and orange colors respectively.
    The 5 reconstruction results are sorted according to the F-Score@0.01 w.r.t the ground truth.}
    \label{fig:test5}
\end{figure}

\noindent\textbf{Performance on Real-World Dataset Pix3D.}
In Tab.~\ref{tab:pix3d1} and Tab.~\ref{tab:pix3d2}, we compare CCD-3DR with other DDPM-based reconstruction methods using Chamfer Distance and F-Score@0.01 respectively.
Our method consistently outperforms competitors in all categories.
On average, CCD-3DR surpasses second-best method PC$^2$ by $20\%$ on ShapeNet-R2N2 and $15\%$ on Pix3D.

\noindent\textbf{Qualitative Comparisons.}
We provide visualization comparisons with previous methods in Fig.~\ref{fig:exp_shapenet} and Fig.~\ref{fig:exp_pix3d}.
It can be seen clearly that our method surpasses competitors with respect to the reconstruction quality.
Particularly, due to our consistent feature conditioning scheme, our method showcases superiority in recovering fine-grained structures, like the hand in the last row of Tab.~\ref{fig:exp_shapenet}.

\begin{table}[]
\begin{center}
\begin{tabular}{ccccc}
\hline
Method            & Chair          & Table     & Sofa  & Average\\
\hline
DMPGen$^*$~\cite{luo2021diffusion} & 0.188 & 0.176 & 0.243 & 0.202
 \\
PC$^2$~\cite{melas2023pc2} &0.336  &0.294 & 0.377 & 0.336
 \\
Ours        & \textbf{0.439} & \textbf{0.559} & \textbf{0.489} & \textbf{0.496}\\
\hline
\end{tabular}
\end{center}
\vspace{-0.5cm}
\caption{\textbf{Performance on Pix3d.} F-Score@0.01 is reported in the table.
Our method outperforms diffusion-based competitors.}
\label{tab:pix3d1}
\end{table}

\begin{table}[]
\begin{center}
\begin{tabular}{ccccc}
\hline
Method            & Chair          & Table     & Sofa  & Average\\
\hline
DMPGen$^*$~\cite{luo2021diffusion} & 53.30 & 50.56 & 21.04 & 41.63
 \\
PC$^2$~\cite{melas2023pc2} &33.21  &13.13  & 3.760 & 16.70
 \\
Ours        & \textbf{14.98} & \textbf{1.475} & \textbf{0.712} & \textbf{5.722}\\
\hline
\end{tabular}
\end{center}
\vspace{-0.5cm}
\caption{\textbf{Performance on Pix3d.}
Chamfer Distance ($\times 10^{-3}$) is reported in this table.
Our method surpasses baselines by a large margin.}
\label{tab:pix3d2}
\end{table}
\begin{table}[]
\begin{center}
\begin{tabular}{c|cccc}
\hline
Occ. Ratio & Method            & Chair          & Table     & Sofa \\
\hline
\multirow{2}{*}{$\sim20\%$} & PC$^2$~\cite{melas2023pc2} &0.324  &0.280 & 0.365
 \\
& Ours        & \textbf{0.424} & \textbf{0.535} & \textbf{0.421} \\
\hline
\multirow{2}{*}{$\sim50\%$} & PC$^2$~\cite{melas2023pc2} &0.310  &0.260 & 0.337
 \\
& Ours        & \textbf{0.411} & \textbf{0.520} & \textbf{0.397} \\
\hline
\end{tabular}
\end{center}
\vspace{-0.5cm}
\caption{\textbf{Ablation studies of robustness towards occlusions.}
Occ. Ratio refers to occlusion ratio.
We report the F-Score@0.01 after randomly masking about 20\% and 50\% visible pixels of the image.}
\label{tab:occ}
\end{table}
\begin{table}[t!]
\centering
\scalebox{0.99}{
\begin{tabular}{cccc}
\shline
Category & Ours-G &
Ours-(G+L) & Ours  \\
\hline
airplane & 0.599 & 0.727 & 0.725  \\
bench & 0.298 & 0.463 & 0.480  \\
cabinet & 0.204 & 0.277  & 0.282 \\
car & 0.251 & 0.398 & 0.395  \\
chair & 0.283 & 0.341 & 0.335  \\
display & 0.223 & 0.366 & 0.381  \\
lamp & 0.316 & 0.429 & 0.438  \\
loudspeaker & 0.177 & 0.214 & 0.219  \\
rifle & 0.653 & 0.777 & 0.762  \\
sofa & 0.201 & 0.287 & 0.293  \\
table & 0.266 & 0.433 & 0.427  \\
telephone & 0.355 & 0.414 & 0.423  \\
watercraft & 0.311 & 0.469 & 0.475  \\
\hline
Average & 0.318 & 0.430 & 0.433  \\
\shline 
\end{tabular} }
\vspace{-0.3cm}
\caption{\textbf{Ablations on the effect of local and global features on ShapeNet-R2N2.} 
We retrain and re-evaluate our method using different feature conditioning methods.
}
\label{tab-gl_ab}
\end{table}

\subsection{Ablation Studies}
We conduct several ablation studies on the public datasets.
Note that except for ablated terms, we leave all other terms and settings unchanged.

\noindent\textbf{Occlusions.}
In Tab.~\ref{tab:occ}, we evaluate the performance of CCD-3DR with respect to different occlusion ratios on Pix3D.
We randomly mask approximately $20\%$ and $50\%$ visible pixels of the object to test the robustness of CCD-3DR towards occlusions.
From the table, it can be seen clearly that although the masked pixels increases from $20\%$ to $50\%$, the performance of CCD-3DR only degrades very little, with 0.013 in \textit{chair}, 0.015 in \textit{table} and 0.024 in \textit{sofa}.
Moreover, in this experiment, PC$^{2}$ also demonstrates consistent and satisfactory results under different occlusion ratios, which verifies the capability of diffusion models in handling occlusions.
Note that for fair comparisons, we retrain PC$^{2}$ and our method by with the same augmented training data.
We randomly mask 0\% $\sim$ 50\% pixels of each image for training and then conduct the ablation study in Tab.~\ref{tab:occ}.

\noindent\textbf{Local \textit{vs.} Global Conditioning.}
In Tab.~\ref{tab-gl_ab}, we demonstrate the effect of local and global features in the diffusion-based reconstruction process.
Global feature is obtained by averaging pooling of the point-wise local features.
And when the global feature is incorporated, we directly concatenate it to each point as the condition.
Comparing \textit{Ours-(G+L)} and \textit{Ours}, it can be seen clearly that once local feature is provided, additional global feature is not necessary. 

\noindent\textbf{Oracle Results.}
We report the oracle experiment results in Tab.~\ref{tab-shapenet}.
Following the setting in PC$^{2}$, we also predict 5 possible shapes for each image and select the one with highest F-Score@0.01 as the final reconstruction result.
It is obvious that under the oracle setting, all the three diffusion-based methods DMPGen$^{*}$, PC$^{2}$ and Ours showcase a significant leap forward.
Thereby, although the centralization scheme in our method may influence the generalization capability of the diffusion model to a certain extent, in the single-view reconstruction case, our method still demonstrate the capability of generating multiple plausible results, as also shown in Fig.~\ref{fig:test5}.   

\section{Conclusions}
In this paper, we propose CCD-3DR, a single-image 3D reconstruction pipeline which leverages a novel centered diffusion probabilistic model for consistent and stable local feature conditioning.
We simply project the predicted noise and sampled point cloud from DDPM into a subspace where the point cloud center remains unchanged in the whole diffusion and reverse processes.
Extensive experimental results and ablation studies on both synthetic and real-world datasets demonstrate that such a simple design significantly improves the overall performance.
We also analyze the influence of point cloud centralization with respect to diversity and point out the limitations of CCD-3DR.
In the future, we plan to extend CCD-3DR with an advanced ordinary differentiable equation solver to enhance the inference speed.

%%%%%%%%% REFERENCES
{\small
\bibliographystyle{ieee_fullname}
\bibliography{egbib}
}

\end{document}